\documentclass[sigconf]{acmart}
\AtBeginDocument{%
  \providecommand\BibTeX{{%
    \normalfont B\kern-0.5em{\scshape i\kern-0.25em b}\kern-0.8em\TeX}}}

\acmConference[XAI-FIN-2022]{2nd International Workshop on Explainable AI in Finance}{November 02--04, 2022}{New-York, USA}
\acmBooktitle{}
\acmPrice{15.00}
\acmISBN{}

\usepackage[english]{babel}
\usepackage[autostyle, english=american]{csquotes}
\MakeOuterQuote{"}
\addto\extrasenglish{
}

\usepackage{makecell}
\usepackage{multirow}
\begin{document}

\title{Explaining Anomalies using Denoising Autoencoders for Financial Tabular Data}
\author{Timur Sattarov}
\affiliation{%
  \institution{Deutsche Bundesbank}
  \city{Frankfurt am Main}
  \country{Germany}}
  \email{timur.sattarov@bundesbank.de}

\author{Dayananda Herurkar}
\affiliation{%
  \institution{German Research Center for Artificial Intelligence (DFKI)}
  \city{Kaiserslautern}
  \country{Germany}}
  \email{dayananda.herurkar@dfki.de}

\author{Jörn Hees}
\affiliation{%
  \institution{Bonn-Rhein-Sieg University of Applied Sciences}
  \city{St. Augustin}
  \country{Germany}}

\affiliation{%
  \vspace{3pt}
  \institution{German Research Center for Artificial Intelligence (DFKI)}
  \city{Kaiserslautern}
  \country{Germany}}
\email{joern.hees@dfki.de}

\renewcommand{\shortauthors}{Sattarov, et al.}

\begin{abstract}
Recent advances in Explainable AI (XAI) increased the demand for deployment of safe and interpretable AI models in various industry sectors.
Despite the latest success of deep neural networks in a variety of domains, understanding the decision-making process of such complex models still remains a challenging task for domain experts.
Especially in the financial domain, merely pointing to an anomaly composed of often hundreds of mixed type columns, has limited value for experts.

Hence, in this paper, we propose a framework for explaining anomalies using denoising autoencoders designed for mixed type tabular data. We specifically focus our technique on anomalies that are erroneous observations. This is achieved by localizing individual sample columns (cells) with potential errors and assigning corresponding confidence scores. In addition, the model provides the expected cell value estimates to fix the errors.  

We evaluate our approach based on three standard public tabular datasets (Credit Default, Adult, IEEE Fraud) and one proprietary dataset (Holdings).
We find that denoising autoencoders applied to this task already outperform other approaches in the cell error detection rates as well as in the expected value rates.
Additionally, we analyze how a specialized loss designed for cell error detection can further improve these metrics. Our framework is designed for a domain expert to understand abnormal characteristics of an anomaly, as well as to improve in-house data quality management processes.

\end{abstract}



\maketitle

\section{Introduction}
\label{sec:intro}

Financial regulatory authorities and supervisory agencies play one of the most important roles in the financial system of a country.
The main objective of the authorities is to secure the financial and monetary stability, supervision of national credit institutions as well as the management of payment service mechanisms.
To fulfill these objectives, national statistical offices of the regulatory authorities need to collect monetary, financial and external sector statistical data.
After the Global Financial Crisis (GFC) of $2008-2009$ the enhancement of the financial framework has become compelling\footnote{https://ec.europa.eu/commission/presscorner/detail/de/MEMO\_13\_679}. In addition to the stronger oversight of financial firms, the GFC led to the call for strengthening and extension of the financial statistics\footnote{https://www.imf.org/external/np/g20/pdf/102909.pdf}. 
Following the above-mentioned initiatives, the demand for high quality financial microdata has appeared.
To monitor the vulnerability of the economy to shocks and identify systemic risks, collection of high-quality microdata plays a vital role.
For National Competent Authorities (NCA), the correctness and completeness of the collected data has to be ensured. Moreover, given the large volumes of collected data today, NCAs have to develop and deploy efficient data quality check (QC) procedures.
Hence, typically a set of handcrafted rules are developed as rudimentary hard-coded checks. However, these are only able to detect already known reported errors and are not capable of identifying new types of errors. 
Further, it is crucial to not only identify an anomalous observation, but also flag the field(s) that contain reporting error(s). Therefore, explaining which values caused an irregularity is essential for financial microdata.

Today, a number of deep learning based techniques are introduced for anomaly detection in tabular data \citep{deep_learning_AD_survey}.
However, in practice such tools are often insufficient due to the lack of interpretation.
The ability to explain anomaly characteristics is as important as the quality of the trained model.
For a domain expert, it is crucial to obtain a comprehensive explanation that would build a connection between a high anomaly score and a set of features affecting this score. Moreover, an inquiry to the reporting agent about the erroneous observation can be made and help with the correction.
Therefore, the utilization of the anomaly interpretation features would significantly improve the applicability of such models in regulatory practice.  

\begin{figure*}[ht]
    \center
    \includegraphics[width=\linewidth]{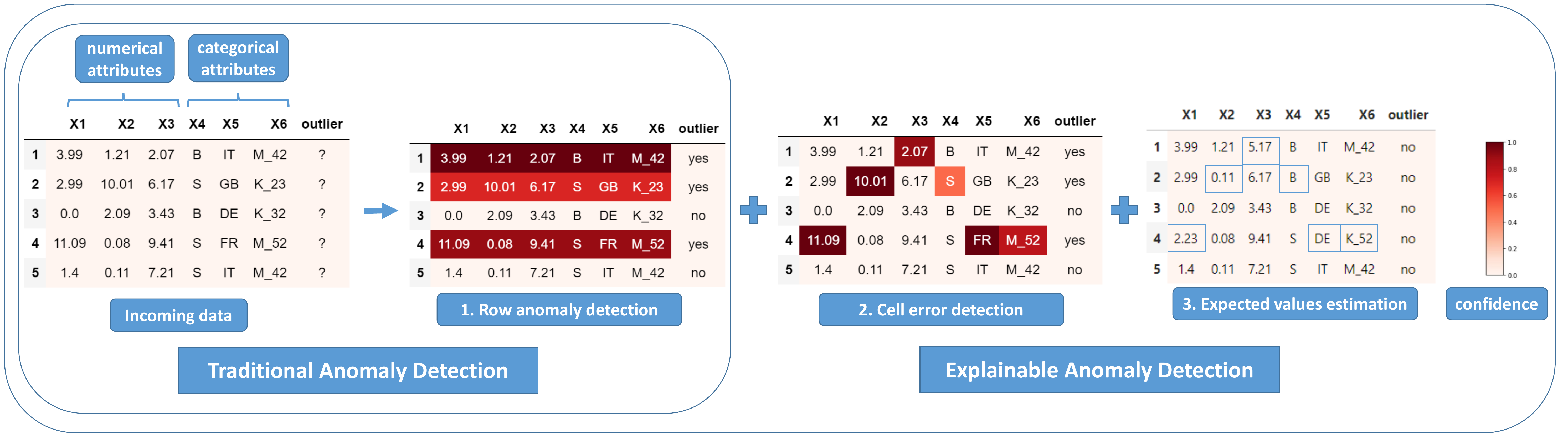}
    \caption{A schematic process overview of explainable anomaly detection (AD) for mixed type tabular data. In comparison to traditional AD which allows only row anomaly detection (1), explainable AD supplies the detection of cells responsible for high anomaly score (2) as well as the estimation of expected values for fixing an error (3). The coloring reflects the error confidence of a particular cell entry.}
    \label{fig:01_sketch_cell_anomaly_score}
\end{figure*}

In this work, we propose a practical framework using denoising autoencoder (DAE) neural networks that not only isolates anomalous data points, but also flags the fields that caused the irregularity. The framework is designed for financial tabular data with categorical and numerical (mixed) type.
\autoref{fig:01_sketch_cell_anomaly_score} illustrates an example of traditional and explainable anomaly detection on financial tabular data. Traditional anomaly detection techniques flag the entire record as an anomaly (step 1) providing only a single score for each observation. This information is not enough to understand the cause of irregularity and only answers the question \textit{"which samples are anomalies?"}. Our framework extends it to explainable anomaly detection providing cell error detection mechanism (step 2) which allows answering the question \textit{"why is it an anomaly?"}. In addition, the model is capable of estimating the expected values, that should have been in place of the errors (step 3). This property allows answering the question \textit{"what should have been reported instead?"}. 
These steps are utilized as the explainability properties of the model and help the domain expert to understand the anomalous characteristics of the detected anomalies.

In summary, we present the following contributions:

\begin{itemize}
    \item We demonstrate that denoising autoencoder neural networks can be utilized to explain the cause of irregularity of a particular sample for mixed type tabular data.
    
    \item We show that such a model can successfully detect reporting errors on the attribute level (cell) providing corresponding confidence scores, as well as proposing the expected estimates for fixing the error.
    
    \item We propose an extension of the model with an enhanced loss and illustrate that such technique outperforms traditional methods based on the selected metrics.
\end{itemize}

The remainder of this paper is structured as follows: \autoref{sec:related_work} provides an overview of the related work. In \autoref{sec:methodology} we describe the autoencoder neural network model with its denoising extension together with the proposed methodology for detecting the erroneous cells. Next, \autoref{sec:setup} and \ref{sec:results} outline the experimental setup and results. We conclude the paper with a summary and future research directions in \autoref{sec:conclusion}.

\section{Related Work}
\label{sec:related_work}

The literature survey hereafter focuses on (1) developed row and cell anomaly detection techniques for financial tabular data, and (2) existing models designed for explainable anomaly detection.

\subsection{Anomaly Detection in Financial Tabular Data}

Anomaly detection has been an active research area in different domains, with a number of methods developed using deep learning \citep{deep_learning_AD_survey}. Especially, tabular data is becoming more and more attractive for deep learning techniques \citep{deep_learning_tabular_survey}.
Nowadays, autoencoders have been widely used not only for representation learning but also for anomaly detection in variety types of financial data \citep{chalapathy2019deep}. Recently, a number of techniques were developed using autoencoders to detect anomalies in large scale accounting data \citep{schreyer2018detection, schreyer2019detection, schultz2020autoencoder}, identify traces of money laundering and fraud \citep{7838276, 8324876} or learn behavioral fraud features \citep{wedge2017solving}. Besides this, Schreyer et al. \citep{schreyer_downstream} have demonstrated successful detection of accounting anomalies in a self-supervised learning setup together with downstream audit tasks. Moreover, autoencoders are a popular technique for detecting credit card fraud schemes \citep{Kazemi2017UsingDN, Pumsirirat2018}. In the context of financial fraud, a number of unsupervised and semi-supervised techniques are gaining popularity \citep{fraud_survey}. 

Recently, Nazabal et al. \citep{nazabal2020handling}  proposed a framework to model variational autoencoders for fitting missing cells in the data. The technique includes handling not only categorical and numerical data types but also ordinal, interval and count. Also, similar to our approach, Eduardo et al. in \citep{eduardo2020robust} proposed the robust version of the VAE for cell-wise outlier detection for mixed type data.

\subsection{Explainable Anomaly Detection}

The field of "Explainable AI" (XAI) is rapidly developing, enhancing variety of the models which help the domain experts slightly open the "black-box" and understand the underlying decision-making process of the complex algorithms \citep{das2020opportunities}. Recently, there have been a number of techniques  introduced \citep{xai_methods2, xia_methods1} in the area of XAI. Such model agnostic methods like SHapley Additive exPlanations (SHAP) \citep{NIPS2017_7062}, Local Interpretable Model-Agnostic Explanations (LIME) \citep{DBLP:journals/corr/RibeiroSG16} or DeepLIFT \citep{DBLP:journals/corr/ShrikumarGK17} showed significant success for their abilities to explain the output of almost any machine learning model.
At the same time, the usage of shapley values is becoming popular in explaining anomalies. Antwarg et al. \citep{antwarg2019explaining} used the kernel SHAP to explain the anomalies detected by the autoencoder neural network in an unsupervised scenario. Similarly,  Takeshi et al. \citep{takeishi2020shapley} successfully used the power of shapley values in linear models such as PCA. 
Nguyen et al. \citep{nguyen2018scalable} have proposed the combined version of the autoencoder and OC-SVM to explain the decision-making process of detected outliers in unsupervised anomaly detection tasks.
 Another unsupervised attempt was made by Chen et al. \citep{Chen2021UnsupervisedAD} to localize structural and non-structural anomalies in computer vision. Previously, Bergmann et al. \citep{Bergmann_2019} proposed the perceptual loss for autoencoders to identify inter-dependencies between local image regions.
Recently, Amarasinghe et al. \citep{xai_anomaly} developed a framework using deep neural networks to explain the cause of detected DoS attacks in a supervised manner. A gradient-based approach was utilized by Nguyen et al. \citep{DBLP:journals/corr/abs-1903-06661} to develop a framework for detecting anomalies in a network traffic using variational autoencoder. Another attempts using attention learning mechanism were proposed by  Venkataramanan et al. \citep{xai_anomaly_attention} and Xu et al. \citep{xai_anomaly_attention2}. 
Also, an explainable recommendation system using autoencoder was developed by  Haghighi et al. \citep{haghighi2019explainable}. The model was designed to explain the outputs of the recommender. Kauffmann et al. \citep{Kauffmann_2020}  used a deep Taylor decomposition to explain various anomaly types. Another practical application to explain the output of black-box model was described by Ramamurthy et al. \citep{ramamurthy2020model}. They build a multilevel explanation tree that characterizes the local and global explanations of the records.
A number of attempts were also made to model the detection of cell errors in medical and geoscience domains.
Jan et al. \citep{cellOD_med1:2019} have proposed a cell outlier diagnostics detection technique and evaluated it on three different medical datasets.
Similarly, the importance of multivariate outlier detection in the field of geosciences was recently demonstrated in  by Filzmoser et al. \citep{TUW-290136}. 

\vspace{0.5em}

According to the systematic review of Riyanul et al. \citep{xai_review_riyanul} only $2 \%$ of the XIA research papers are focused on the finance domain.
The literature survey above also demonstrated the overall popularity of XAI techniques, but very limited application of anomaly explanations for financial data, especially in combination with denoising autoencoder neural networks. 

\section{Methodology}
\label{sec:methodology}

In this section, we describe the autoencoder neural network, its denoising extension with the proposed loss, as well as the specification of the framework for explaining anomalies.

\subsection{Autoencoder Neural Network}
Formally, we denote a set of instances $x_1,x_2,...,x_N$ in a tabular dataset $X$. Every instance encompasses a set of attributes $d\in{\{1,...,D\}}$ with either numeric $x_{n}^{d_\text{num}}\in{\mathbb{R}}$ or categorical type $x_{n}^{d_\text{cat}}\in{\{1,...,C\}}$ where $C$ is the total number of unique categories of the feature $d$.

An autoencoder (AE) neural network is a type of feed-forward network that aims to perform a lossy data compression into a lower dimensional feature space and afterwards reconstruct it back to the original data space with minimal loss.
The encoder network $f_\theta$ performs the data compression and the decoder network $g_\psi$ accomplishes the reconstruction.
Upon the successful model training with a set of parameters $\theta$ and $\psi$, the reconstruction error is often used to quantify the anomaly degree of an instance. The reconstruction error reflects how good an instance fits into the general patterns of the data. Hence, an inlier receives a relatively low reconstruction error, whereas an outlier obtains a higher one, which attests its deviation from the common data structure.
The network is trained in an end-to-end unsupervised fashion by minimizing the reconstruction loss, formally defined as follows:

\begin{equation}
	\arg\min_{\theta \psi} \|X - g_\psi(f_\theta(X))\|
	\label{equ:ae_objective}
\end{equation}

Due to the mixed type nature of the data, we define the reconstruction loss of every instance as the sum of two losses.
For each one-hot encoded representation of the categorical attribute the (1) negative-log-likelihood loss is calculated and (2) the mean-squared loss used for the numerical attributes, formally expressed by:

\begin{equation}
    \mathcal{L}_{\theta,\psi}(x^{d}_n;\hat{x}^{d}_n) =   \sum_{d=1}^{D_\text{cat}} \mathcal{L}^\text{NLL}_{\theta,\psi}(x^d_n;\hat{x}^d_n) +  \sum_{d=1}^{D_\text{num}}
    \mathcal{L}^\text{MSE}_{\theta,\psi}(x^d_n;\hat{x}^d_n)
    \vspace{2mm}
    \label{equ:reconstruction_loss_details}
\end{equation}

\noindent where $\hat{x}^{d}_n$ denotes the $n-th$ reconstructed sample and its attribute $d$. We have observed that such loss design suits better for mixed data type as it leads to a faster overall model convergence.

\subsection{Denoising Autoencoder Neural Network}
\label{sec:DAENN}

The denoising autoencoder (DAE) is an extension of the traditional autoencoder neural network with the goal of removing noise from the signal. 
Such a model is trained by disrupting the input data with random noise and reconstructing the clean data. 
At first, a corrupted instance $\tilde{x}$ is created by adding random noise to the clean input instance $x$.  Next, the encoder network $f_\theta$ performs the compression of the corrupted instance $\tilde{x}$, and the decoder network $g_\psi$ accomplishes the reconstruction $\hat{x}$ (noise removal). The training objective function is the same as in the \autoref{equ:ae_objective}. In the inference phase, the trained model is capable of transforming the noisy data into noiseless.
In addition to the denoising capabilities, such model modification improves the robustness of the hidden layers (i.e., latent layer representation) \citep{Vincent2008} as well as reduces the risk of overfitting.

\begin{figure*}[th]
    \center
    \includegraphics[width=\textwidth]{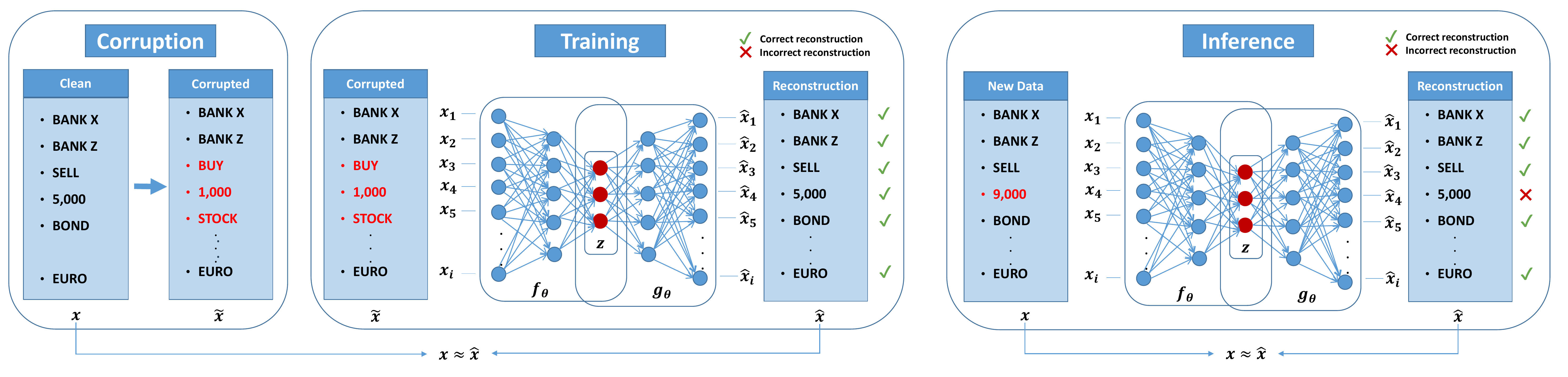}
    \caption{A schematic overview of the training and inference phases of DAE  for cell error detection and estimation of expected values. In the corruption phase, random noise is added to a sample. Next, the DAE is trained on the corrupted data with the goal to reconstruct the clean data. In the inference phase, the cells that failed to reconstruct, are flagged as an error, and the reconstructed value is used as an estimation of the expected value to fix the error.}
    \label{fig:02_sketch_DAE_overview}
\end{figure*}

\noindent \textbf{Enhanced loss.} We have noticed, that the selection of the right amount of injected noise during training is quite challenging for mixed type tabular data. Too much noise leads the model to focus mainly on the noise removal task, and the network fails to reconstruct clean data sufficiently enough. On the contrary, too little noise decreases noise removal capabilities, which affects the overall cell error detection rate.
Therefore, we propose an extension to the loss function of the DAE.
Specifically, we introduce a parameter $\alpha=[0,1]$ that allows us to weigh the noise removal vs. clean data reconstruction within the batch.
In practice, we found that selection of a fixed $\alpha$ is challenging.
Therefore, we propose to sample it from the \textit{Beta(0.5, 0.5)} distribution. 
The random sampling of $\alpha$ can be understood as a regularization technique (somewhat similar to dropout) to optimize both of our goals (noise removal and reconstruction of clean data).
The alternating nature of $\alpha$ (i.e., being sampled close to the extremes of $0$ or $1$) seems to be beneficial to training in many cases.
The final loss has the following formulation:

\begin{equation}
    \mathcal{L}_{\theta,\psi}(x^{d}_n;\hat{x}^{d}_n) =  
    \alpha \hspace{1mm} \mathbf{m} \odot \mathcal{L}_{\theta,\psi}(x^{d}_n;\hat{x}^{d}_n) + (1-\alpha) \hspace{1mm} \mathbf{\bar{m}} \odot \mathcal{L}_{\theta,\psi}(x^{d}_n;\hat{x}^{d}_n)
    \vspace{2mm}
    \label{equ:noise_balancer}
\end{equation}

\noindent where $\mathbf{m}$ is a binary mask vector  $\mathbf{m}\in \{0,1\}^d$ that yields $1$ at the entry with noise or $0$ otherwise, $\mathbf{\bar{m}}$ is its complement, and $\odot$ is the element-wise multiplication.

\subsection{Anomaly Explainer}
\label{subsec: anomaly explainer}
To explain the cause of an anomaly, we utilize the properties of the reconstruction error of each separate attribute of the trained DAE. 
Such an approach is also quite common in practice using traditional AE. Although this typically yields good performance on the detection of 
row anomalies, it becomes less precise in identifying the exact cells that contain errors. Hence, the goal of the proposed framework is to answer three questions by supplying the domain expert with the following information:

\begin{itemize}
    \item \textbf{Which samples are anomalies?} \textit{Row Anomalies}:  identify a subset of $K$ row anomalies with the highest reconstruction errors.
    \item  \textbf{Why is it an anomaly?} \textit{Cell Errors}: for every cell in $K$ selected anomalies, compute the confidence $\pi_n^d$ that the value $x_n^d$ contains an error. 
    \item \textbf{What should have been reported instead?} \textit{Expected Values}:  for every cell in $K$ selected anomalies, collect the reconstructed value $\hat{x}^d_n$. 
\end{itemize}

\noindent \textbf{Training.} As depicted in \autoref{fig:02_sketch_DAE_overview} the DAE is trained to reconstruct a clean (noiseless) instance $x_n$ from its corrupted counterpart $\tilde{x}_n$. During the training, the reconstruction error between the clean instance $x_n$ and its reconstruction $\hat{x}_n$ is minimized.

\noindent \textbf{Inference.} Once the DAE is trained, the reconstruction error $\hat{x}^d_n$ of each attribute value of the test (unseen) data is calculated. Depending on the attribute type (either categorical or numerical) two different functions are applied to obtain the error confidence $\pi_n^d$ of this cell.
\noindent For categorical attributes, we compute the 
complement of the normalized reconstruction category $c$ as the following:

\begin{equation}
    \textbf{Cell: } \pi^{d_\text{cat}}_n =  1 - a^{dc}_n
    \label{equ:prob_of_error_cat}
\end{equation}

\noindent where $a(\cdot)$ is the softmax function $a^{dc}_n=\text{softmax}(\hat{x}^d_n)^c$ calculated on the reconstructed representation of the attribute $d$. The superscript $c$ identifies a particular category in that attribute.
For numerical attributes, we compute the complement of the negative exponential function between the input value $x^d_n$ and its reconstruction $\hat{x}^d_n$ as the following:

\begin{equation}
    \textbf{Cell: } \pi^{d_\text{num}}_n = 1 - e^{-(x^{d}_n - \hat{x}^d_n)^2 }
    \label{equ:prob_of_error_num}
\end{equation}

\noindent Correspondingly, the row anomaly score is computed as the sum of all categorical and numerical cell scores $\pi_n^d$:

\begin{equation}
    \textbf{Row: } \pi_n = \sum_{d=1}^D \pi^d_n 
    \label{equ:prob_of_anomaly}
\end{equation}

\noindent The expected values are obtained by collecting the reconstructed values $\hat{x}_n^d$. For categorical attributes, we use the highest probability category $\arg\max_{c}$  $a_n^{dc}$ of the softmax transformation $a_n^d$.



\section{Experimental setup}
\label{sec:setup}

In this section, we describe the details of the conducted experiments.
We describe the datasets as well as the noise injection procedure that was applied to these datasets, together with the metrics used to evaluate the performance of the results.
For training and evaluation of the neural network models, the PyTorch v$1.10.2$ \cite{pytorch} framework was used.

\subsection{Datasets}

We benchmark the developed technique with open-source and real world datasets. 
Three public datasets and one proprietary dataset were selected to evaluate the performance of the proposed framework.
Below, we provide the description of each dataset: 

\begin{itemize}
    \item \textbf{Credit Default}\footnote{The dataset is publicly available via: \url{https://archive.ics.uci.edu/ml/datasets/default+of+credit+card+clients}}: The dataset is taken from the UCI machine learning repository \citep{Dua:2019} and contains information on bill statements of credit card clients, their default payments, history of payment as well as the demographic factors of the clients in Taiwan during the period April 2005 to September 2005 \citep{credit_default}.
    
    \item \textbf{IEEE Fraud}\footnote{The dataset is publicly available via \url{https://www.kaggle.com/c/ieee-fraud-detection/overview}}: The dataset consists of the electronic transactions from the e-commerce service provider Vesta Corporation. The dataset was published to improve the efficiency of the fraud detection alert system.
    
    \item \textbf{Adult}\footnote{The dataset is publicly available via \url{https://archive.ics.uci.edu/ml/datasets/adult}}: The dataset is taken from the UCI machine learning repository \citep{Dua:2019} and consists of personal income records, where the task is to predict whether an income exceeds \$$50$k per year. 
    
    \item \textbf{Holdings}\footnote{In compliance with strict data privacy regulations, neither content nor the descriptive statistics of the dataset can be made publicly available.}: This proprietary dataset consists of the individual holdings of the investment funds issued by investment companies \citep{Holdings}. Each record reflects the asset or liability value submitted by the reporting entity at the end of the month. 
\end{itemize}

All datasets have mixed attribute types as described in \autoref{tab:datasets}. In the data preprocessing step, all categorical attributes are encoded using the one-hot encoding technique. Numerical attributes are standardized to have $0$ mean and standard deviation $1$.
Afterwards, the one-hot encoded representation is combined with standardized numerical attributes. The final number of encoded attributes is reflected in the last column ("Encoded") of the Table \ref{tab:datasets}.

\begin{table}[h]
\caption{\label{tab:datasets} Descriptive statistics of the selected datasets}
\begin{tabular}{lrrrr}
\toprule
\multirow{2}{*}{Data}  &  \multirow{2}{*}{Rows} & \multicolumn{3}{c}{Columns}  \\
                        &                     & Categ. & Num.  & Encoded \\

\midrule
Credit Default  & 30,000    & 10   & 13 & 160 \\
IEEE Fraud      & 569,877   & 14  & 380 & 502 \\
Adult           & 32,561    & 8   & 5 & 126 \\     
Holdings       & 118,569   & 7   & 129 & 203 \\
\bottomrule
\end{tabular}
\end{table}


\subsection{Corruption process} 


To the best of our knowledge, there is no publicly available dataset with labeled cell errors.
Therefore, it is a standard practice to artificially generate anomalies by randomly corrupting the clean data \citep{cell_corrupt_redyuk, cell_corrupt_krishnan}. 
In our approach, we also follow a similar strategy and turn $3\%$ of the inliers into the outliers by randomly corrupting attribute values in both training and test sets.
Selection of the attributes for data corruption is also done at random.
We corrupt at most $50\%$ of the features which are selected uniformly at random as following: $c = \text{Unif}(1, \tfrac{c_{max}}{2})$, where $c_{max}$ is the total number of features. 
Dataset Holdings already contains the real-world cell errors together with its clean value.

To artificially corrupt the samples, we applied different techniques for both categorical and numerical features.

\noindent \textbf{Numerical features.}
The injection of noise for a numerical feature is performed using an additive noise process, with the corrupted value obtained as: $\tilde{x}_{n}^{d} = x_{n}^d + \delta $. Here $\delta$ is randomly sampled from one of the Gaussian, Laplace, or Log-Normal  distributions with $\mu=0$ and $\sigma=\sigma_d \gamma $. Selection of $\gamma$ follows uniform distribution $\gamma=\text{Unif}(3,5)$ and $\sigma_d$ is the standard deviation of the original attribute. The selection of  the distribution at the corruption phase is also done uniformly at random.

\noindent \textbf{Categorical features.}
Two alternatives are used to inject a noise into categorical attributes. With the first alternative, the original entry is replaced by picking a categorical entry uniformly at random from the distinct values of this attribute. The second option creates a new categorical entry by performing character manipulations (insertion, flipping or deletion) with the original categorical entry and ensuring a completely new entry is created. Such technique in practice imitates a typo that can often appear during the data insertion process.

    
    
        

\subsection{Evaluation metrics}

To assess the quality of the proposed technique, we utilize the following three metrics and measure the detection rate.

\noindent \textbf{Precision at K (P@K).} We utilized this metric for the traditional row anomaly detection to assess overall model capabilities to detect anomalies. Hence, this metric is referred to our first question, \textit{"which samples are anomalies?"}. The metric is popular in recommendation system evaluation tasks, where the user is interested only in the top $K$ predictions. Similarly, in a regulatory reporting environment, it is important that top $K$ retrieved anomalies are indeed relevant, hence reducing the false positive rate as well as human effort.

\begin{equation}
    \text{P@K}(y, \hat{y}) = \frac{\text{TP@K}(y, \hat{y})}{K}
    \label{eq:mp@k}
\end{equation}

\noindent where $\text{TP@K}$ is the total number of true anomalies in the top $K$ retrieved anomalies given the vectors $y$ and $\hat{y}$ of true anomalies and row-wise reconstruction errors correspondingly. The value of $K$ in our case is selected as the total number of true anomalies in the test set.

\noindent \textbf{Mean Average Precision (mAP).} This metric reflects the performance of the model in answering the second question, \textit{"why is it an anomaly?"}. Thus, it estimates the quality of cell error detection across all attributes. The confidence of the cell error $\pi^d$, described in \autoref{subsec: anomaly explainer}, is used as the input to the function for computing the Average Precision (AP). The positive labels in this case are the cells with noise. Formally, it is defined as: 

\begin{equation}
    \text{AP}(\pi^d) = \sum_{i=1}^{N} (R_i - R_{i-1})P_i
    \label{eq:map}
\end{equation}

\noindent where $R_i(\pi^d) = TP / (TP + FN)$  denotes the detection recall and $P_i(\pi^d) = TP / (TP + FN)$ denotes the detection precision of the $i-th$ anomaly score threshold. The mean Average Precision (mAP) is computed as the average of the AP scores across all attributes $mAP = \frac{1}{D} \sum_{d=1}^{D} AP(\pi^d)$.

\noindent \textbf{Mean Expected Value (mEV).} With this metric, we evaluate the ability of the model to answer the third question, \textit{"what should have been reported instead?"}. In order to assess the correctness of the expected (or fixed) values, we compute the Standardized Mean Squared Error (SMSE) between the original ground truth and its reconstruction.

For numerical attributes, it is additionally normalized by the empirical variance of this attribute and has the following form:

\begin{equation}
    EV(x^{d_{\text{num}}}) = \frac{1}{\hat{N}} \sum_{n=1}^{\hat{N}} \frac{(x^d_{no} - \hat{x}^d_n)^2} {\sigma^2}
    \label{eq:mev_num}
\end{equation}

\noindent where $\sigma$ is the standard deviation and $\hat{N}$ is the total number of corrupted cells in the attribute $x^d$. The subscript $o$ in $x^d_{no}$ denotes the ground truth original value (i.e., without error). 

For categorical features, we utilize the \textit{Brier score} \citep{brier} between the one-hot representation of the ground truth value and the reconstructed softmax representation of this category and has the following form:

\begin{equation}
    EV(x^{d_{\text{cat}}}) = \frac{1}{2\hat{N}} \sum_{n=1}^{\hat{N}} \sum_{c=1}^{C} (x^{dc}_{no} - \hat{x}^{dc}_{n})^2
    \label{eq:mev_cat}
\end{equation}

\noindent where $x^{dc}_{no}$ is the one-hot encoded value of the category $c$. The factor $\frac{1}{2}$ is used to normalize the score to the range $[0,1]$. 

The mean Expected Value (mEV) is computed as the average of Expected Values (EV) $mEV =  \frac{1}{D} \sum_{d=1}^{D} EV(x^d)$ across all attributes.

\subsection{Model training setup}

We split every dataset into training and test sets by a fraction of $70$ and $30$ correspondingly. According to the anomaly injection process described in \autoref{sec:methodology} the test set (and if necessary, train set) is populated with noise. Once the model is trained, all evaluation metrics are collected on the test set. The exact network architecture used for each dataset is described in \autoref{tab:ae_architecture}.

\begin{table}[h]
\caption{Selected architecture setup of the (denoising) autoencoder neural network used for each dataset}
\label{tab:ae_architecture}
\begin{tabular}{lc}
\toprule
Dataset & Neurons per hidden layer  \\
\midrule
Credit Default  & 160-128-64-128-160 \\
Adult           & 126-128-64-128-126 \\
IEEE Fraud      & 502-512-256-512-502 \\
Holdings       & 203-256-128-256-203 \\
\bottomrule
\end{tabular}
\end{table}

We train every model for a maximum of $5000$ epochs with a mini-batch of size $128$ and use the Adam optimizer \cite{kingma2014method} with $\beta_{1}=0.9$, $\beta_{2}=0.999$ in combination with a cosine learning rate scheduler. The parameters of the encoder and decoder are randomly initiated as described in \cite{Glorot10understandingthe}.

\noindent \textbf{Baseline models.} To illustrate the practical applicability of the proposed technique, we compare its performance against several methods where cell error detection is feasible. Therefore, we select PCA \citep{PCA}, \textit{Marginal Distribution} and traditional AE \citep{Hinton:2006}. For \textit{Marginal Distribution,} we follow the same approach described by Eduardo et al. \citep{eduardo2020robust} and fit a Gaussian mixture model on every numeric attribute separately, using the negative log-likelihood as the cell error. For categorical attributes, a normalized category frequency is used for expected value estimation. For the AE, we evaluate two scenarios: the training set contains anomalies (AE with anomalies) and the training set does not contain anomalies (AE no anomalies). 
The first scenario imitates the case in industry when the AE is trained from scratch every time new data arrives, without any knowledge about the historical data. The second scenario imitates the case when the historical data with cell errors and their corrected values is available. Here it is possible to train the model on a pure "clean" version of the historical data and evaluate on the unseen data with anomalies.

\section{Experimental results}
\label{sec:results}

In this section, we describe the results of the conducted experiments. We demonstrate the practical applicability with qualitative assessment as well as the efficiency of the proposed technique, providing the quantitative results.

\subsection{Qualitative evaluation}

\begin{figure*}[ht]
    \centering
    \includegraphics[width=\textwidth]{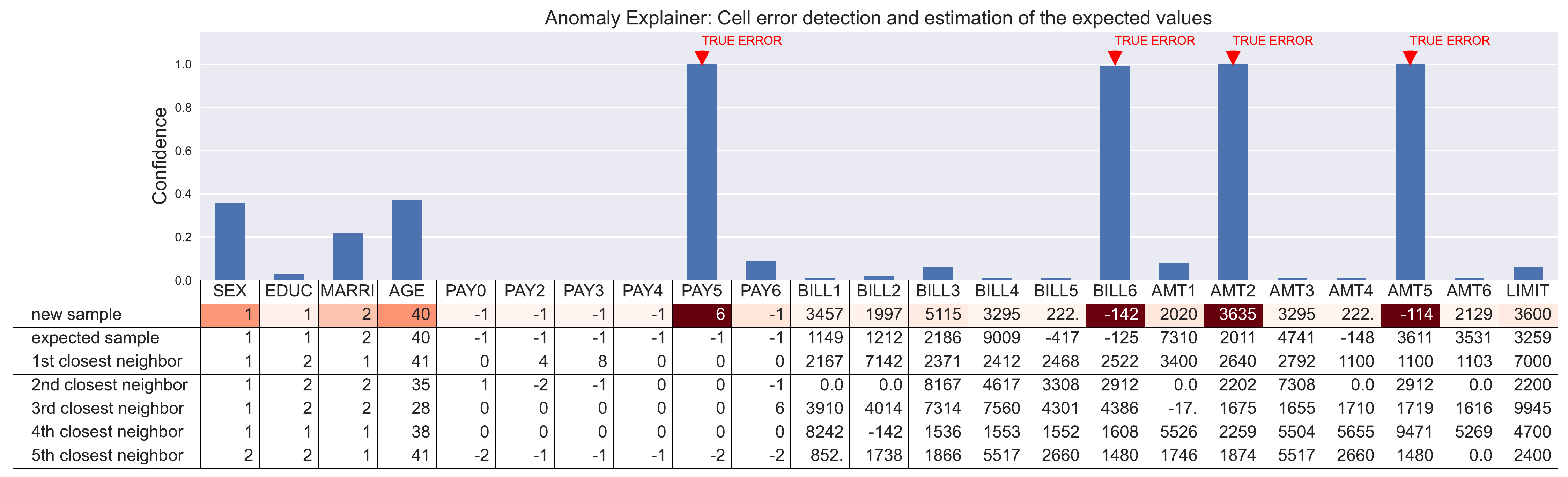}
    \caption{Anomaly explainer dashboard that illustrates the outcome yield by the model trained on the Credit Default dataset. A random anomaly ("new sample") is picked from the test set. Potential cell errors are colored with red gradient and corresponding confidence is reflected in bar graphs. The red arrows point to the position of the true errors. The second row contains the expected values estimated by the model, and the remaining rows are the 5 closest original data instances based on the transformed low dimensional latent representation.}
    \label{fig:qualitative}
\end{figure*}

To explain the cause of irregularity of a potential anomaly, the framework arms the domain expert with a powerful visual inspection tool. Every potential anomaly can be quickly screened and the  question \textit{"why is it an anomaly?"} can be answered. This is achieved by flagging individual cells with detected errors. \autoref{fig:qualitative} depicts the interface of the cell error detection framework. Here, the height of the bars reflects the model's confidence about the reported errors of a new sample. Next, to allow the domain expert to answer the question \textit{"what should have been reported instead?"}, the framework proposes the expected sample. It gives an estimation of the expected values to be reported. In addition to the cell scores, five similar data samples picked from the original dataset are shown under the screened sample. This allows the domain expert to compare certain entries of the screened sample with the entries of its closest neighbors. Selection of such samples is computationally inexpensive, since the pairwise distances are computed on the transformed representation of the bottleneck layer of the DAE. With this tool, the domain expert can pick any data sample, produce such a graphic to quickly assess the nature of the reported errors and execute necessary steps, if required.  Such a compact form (1) provides more explanation capabilities about the anomaly nature, (2) saves the screening time, and (3) reduces the human error during the quality checks.

\begin{figure*}[h]
    \includegraphics[width=\textwidth]{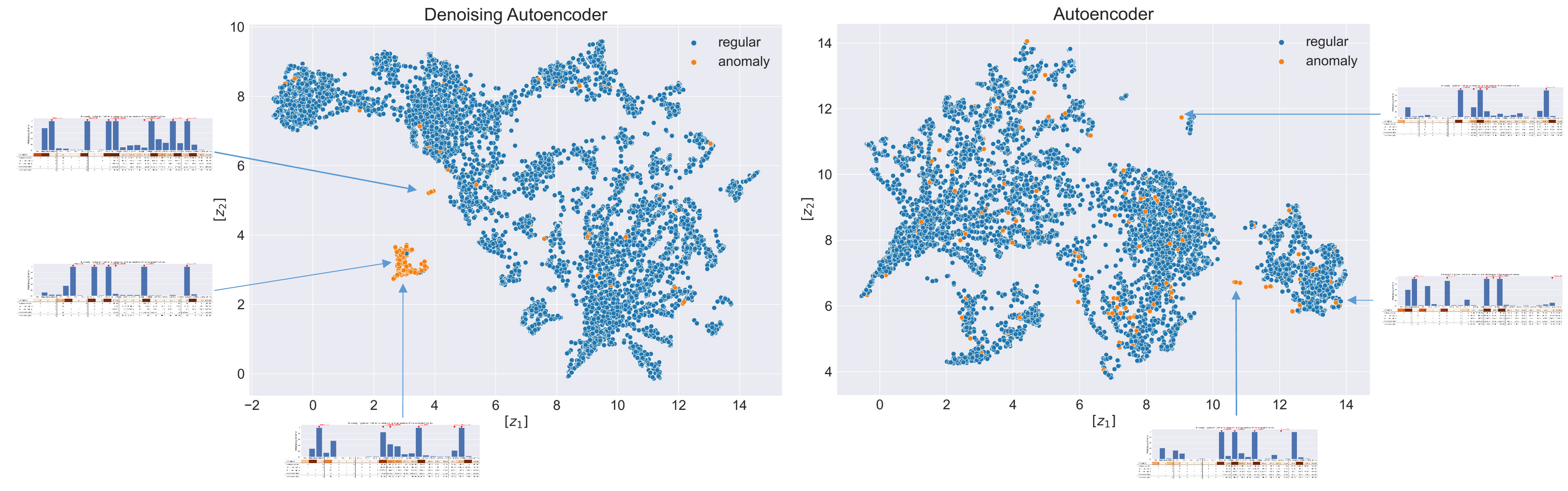}
    \caption{UMAP embedding of the latent representation $z$ between the denoising autoencoder (left) and autoencoder (right) neural networks. The embedding is done on the Credit Default dataset and is projected from 64 to 2 dimensional space of the converged model after 5000 training epochs.}
    \label{fig:umap}
\end{figure*}

\noindent \textbf{Latent space.}
Sampling of the anomalies for screening can also be done using latent data representation produced by the model. This is another powerful property of the framework that allows the domain expert to visually inspect the groups of observations. AEs possess this capability as they are able to learn expressive low dimensional representations of the data in the latent space. Albeit tree based models have the ability to plot the decision tree hierarchy, which makes them indeed preferable tool in industry, they lack the ability to provide the learned data representation with relative similarities between the observations. An example of such data representation is depicted on \autoref{fig:umap}. Here we plot the latent representation of the trained DAE and AE. The former provides a better isolation of anomalies from the regular data points by grouping them into a single cluster. In addition, the DAE seems to provide a more compact form of grouping the regular data points with similar characteristics. In contrast, the AE yields a sparse form of the data representation and anomalies also scattered across the whole latent space. Such representation becomes less valuable for the domain expert who expects to have more compact data representation with more or less clear group separations. This property is especially important in industry because it allows to "walk in the data" and quickly sample the data for screening. The domain expert can sample any data point and produce the anomaly explainer dashboard like on \autoref{fig:qualitative}. This gives an opportunity to quickly audit (financial) entities with similar underlying characteristics (1) as well as the entities that change their cluster assignment (2) which could lead to behavioral changes.

\subsection{Quantitative evaluation}

\begin{table*}[ht!]
\begin{tabular}{ c l | c | c c | c c }
\toprule
\multirow{2}{*}{Dataset} & \multirow{2}{*}{Model} & \multirow{2}{*}{P@K $\uparrow$} & \multicolumn{2}{c}{mAP $\uparrow$}  & \multicolumn{2}{c}{mEV $\downarrow$}  \\
&  &  & categorical & numerical         & categorical & numerical (log)       \\ 
\midrule
\multirow{6}{*}{Credit Default} & PCA & 0.584 $\pm$ 0.004 &  0.709 $\pm$ 0.013 & 0.211 $\pm$ 0.001 & 0.446 $\pm$ 0.001 & 2.750 $\pm$ 0.001  \\
               & Marginals & 0.577 $\pm$ 0.013 & 0.539 $\pm$ 0.000 & 0.444 $\pm$ 0.002 & 0.343 $\pm$ 0.000 & 1.975 $\pm$ 0.022  \\
               & AE with anomalies & 0.651 $\pm$ 0.022 & 0.350 $\pm$ 0.029 & 0.238 $\pm$ 0.012 & 0.821 $\pm$ 0.011 & 2.846 $\pm$ 0.016  \\ 
               & AE no anomalies & 0.814 $\pm$ 0.008 & 0.615 $\pm$ 0.034 & 0.476 $\pm$ 0.009 & 0.633 $\pm$ 0.024 & 2.177 $\pm$ 0.028  \\
               & DAE & 0.826 $\pm$ 0.005 & 0.818 $\pm$ 0.007 & 0.617 $\pm$ 0.005 & 0.245 $\pm$ 0.004 & 0.428 $\pm$ 0.037  \\ 
               & DAE$^*$ & \textbf{0.835} $\pm$ \textbf{0.007} & \textbf{0.821} $\pm$ \textbf{0.006} & \textbf{0.635} $\pm$ \textbf{0.015} & \textbf{0.243} $\pm$ \textbf{0.004} & \textbf{0.415} $\pm$ \textbf{0.035}  \\ 
\midrule
\multirow{6}{*}{Adult} & PCA & 0.328 $\pm$ 0.000 & 0.135 $\pm$ 0.001 & 0.228 $\pm$ 0.001 & 0.422 $\pm$ 0.000 & 2.069 $\pm$ 0.001  \\
               & Marginals & 0.620 $\pm$ 0.004 & 0.192 $\pm$ 0.000 & \textbf{0.626} $\pm$ \textbf{0.008} & 0.299 $\pm$ 0.000 & 1.988 $\pm$ 0.061  \\
               & AE with anomalies & 0.478 $\pm$ 0.007 & 0.144 $\pm$ 0.013 & 0.294 $\pm$ 0.011 & 0.953 $\pm$ 0.005 & 2.966 $\pm$ 0.012  \\ 
               & AE no anomalies & 0.634 $\pm$ 0.014 & 0.262 $\pm$ 0.009 & 0.493 $\pm$ 0.006 & 0.867 $\pm$ 0.012 & 2.536 $\pm$ 0.005  \\
               & DAE & 0.636 $\pm$ 0.015 & \textbf{0.544} $\pm$ \textbf{0.013} & 0.528 $\pm$ 0.013 & 0.451 $\pm$ 0.020 & \textbf{1.572} $\pm$ \textbf{0.043}  \\ 
               & DAE$^*$ & \textbf{0.638} $\pm$ \textbf{0.003} & 0.532 $\pm$ 0.010 & 0.538 $\pm$ 0.007 & \textbf{0.440} $\pm$ \textbf{0.006} & 1.725 $\pm$ 0.035  \\ 
\midrule
\multirow{6}{*}{IEEE Fraud} & PCA & 0.906 $\pm$ 0.001 &  0.622 $\pm$ 0.006 & 0.352 $\pm$ 0.001 & 0.487 $\pm$ 0.001 & 4.554 $\pm$ 0.001  \\
               & Marginals & 0.972 $\pm$ 0.001 &  0.325 $\pm$ 0.000 & \textbf{0.819} $\pm$ \textbf{0.001} & 0.293 $\pm$ 0.000 & 4.474 $\pm$ 0.001  \\
               & AE with anomalies & 0.802 $\pm$ 0.008 &  0.531 $\pm$ 0.005 & 0.265 $\pm$ 0.006 & 0.787 $\pm$ 0.012 & 4.587 $\pm$ 0.032 \\
               & AE no anomalies & \textbf{0.975} $\pm$ \textbf{0.001} &  0.445 $\pm$ 0.014 & 0.510 $\pm$ 0.008 & 0.555 $\pm$ 0.021 & 4.412 $\pm$ 0.102  \\
               & DAE & \textbf{0.975} $\pm$ \textbf{0.004} &  \textbf{0.766} $\pm$ \textbf{0.020} & 0.784 $\pm$ 0.011 & 0.228 $\pm$ 0.004 & \textbf{4.005} $\pm$ \textbf{0.003}  \\
               & DAE$^*$ & 0.974 $\pm$ 0.001 &  0.765 $\pm$ 0.014 & 0.786 $\pm$ 0.006 & \textbf{0.227} $\pm$ \textbf{0.006} & 4.015 $\pm$ 0.125  \\
\midrule

\multirow{6}{*}{Holdings} & PCA & 0.200 $\pm$ 0.007 &  0.005 $\pm$ 0.001 & 0.042 $\pm$ 0.003 & \textbf{0.500} $\pm$ \textbf{0.001} & 13.325 $\pm$ 0.001  \\
               & Marginals & 0.092 $\pm$ 0.002 &  0.001 $\pm$ 0.000 & 0.083 $\pm$ 0.001 & 0.535 $\pm$ 0.000 & 11.610 $\pm$ 0.001  \\
               & AE with anomalies & 0.163 $\pm$ 0.017 &  0.010 $\pm$ 0.010 & 0.040 $\pm$ 0.004 & 0.974 $\pm$ 0.028 & 13.177 $\pm$ 0.100  \\
               & AE no anomalies & 0.157 $\pm$ 0.019 &  0.012 $\pm$ 0.006 & 0.039 $\pm$ 0.003 & 0.942 $\pm$ 0.077 & 13.169 $\pm$ 0.115  \\
               & DAE & \textbf{0.206} $\pm$ \textbf{0.005} &  \textbf{0.045} $\pm$ \textbf{0.010} & \textbf{0.098} $\pm$ \textbf{0.009} & 0.736 $\pm$ 0.035 & \textbf{11.576} $\pm$ \textbf{0.018}  \\
               & DAE$^*$ & 0.201 $\pm$ 0.007 &  0.030 $\pm$ 0.002 & 0.081 $\pm$ 0.006 & 0.709 $\pm$ 0.066 & 11.632 $\pm$ 0.034  \\
\bottomrule
\end{tabular}
\caption{Comparative performance evaluation of the proposed model against the baselines on all datasets using three metrics. The model marked with asterisk was trained using the enhanced loss described in \autoref{sec:DAENN}. Every score reflects the mean and standard deviation from 5 experiments with varying initialization seeds. We can see that DAE outperforms its counterparts on average by 5\%-30\%.}
\label{tab:quantitative_scores}
\end{table*}

We are interested in the precise localization of errors in cells, as this explains the characteristics of an anomaly. As described earlier, we assess the quality of the proposed technique using different metrics, datasets and baseline models. \autoref{tab:quantitative_scores} contains the scores collected from the conducted experiments. Based on these, the DAE outperforms the baseline AEs on almost all metrics and datasets. We believe that this is due to the fact that the traditional AE yields high reconstruction errors not only on the corrupted cells, but on other (neighboring) cells as well, which produces lots of false positives.
Instead, the DAE due to its training nature, reconstructs each cell more precisely and hence, produces less false positives. In the cases where it concedes (mAP of numerical attributes of Adult and IEEE Fraud), we believe the reason lies in the uninformativeness of certain attributes \citep{tree_based_model_better}. We have noticed, that for such scenarios, the Marginal model yields better results.

In addition, the AE trained only on clean data (AE no anomalies) almost always outperforms its counterpart trained on the data with anomalies (AE with anomalies).
We believe, that this happens because the AE trained with anomalies at some point during training shifts its focus towards learning the anomalies since they are responsible for the highest reconstruction errors. As a result, in the inference phase, the anomalies are getting lower reconstruction errors. 
That implies that in practice it's better to deploy the model trained on the "clean" data (if available) rather than to retrain an AE on new (potentially noisy) data from scratch. 
Even more efficient is to use the historical anomalies and let the model learn from this. 

The results on the real world dataset Holdings attest this.
Since the dataset Holdings contained the clean and noise versions, we were also able to compare the performance of the DAE with various noise types. Three models were trained using only artificial noise (1), only real world noise (2) and real world + artificial noise (3). Based on the collected scores, using both (3) boosts the performance notably. This is expected, as it allows the DAE (unlike the AE) to also learn from the distribution of real world noise during training. 

\section{Conclusion and Future work}
\label{sec:conclusion}
In this work, we proposed a framework for explaining detected anomalies using denoising autoencoder neural networks for mixed type tabular data. 
To explain the cause of an anomaly, the framework produces confidence scores of potential errors for every cell entry, as well as proposes corresponding estimated values to fix the errors.
In addition, we propose the enhanced extension using the extended loss specifically designed for cell error detection.

We evaluated the proposed approach on three publicly available datasets and one proprietary financial dataset with mixed type attributes. The produced results are compared against the baseline and underpin the practical applicability of the proposed technique. 

We believe that such a framework can become a helpful toolbox for data quality experts in their daily tasks and can be easily integrated into the corresponding procedural pipeline. We believe the technique can also be applied in a variety of other domains outside the financial field in the future.

\section{Acknowledgments}
We thank the members of the statistics department at the Deutsche Bundesbank for their valuable review and remarks. Opinions expressed in this work are solely those of the authors and do not necessarily reflect the view of the Deutsche Bundesbank or its staff.
Further, we would like to thank Marco Schreyer (University of St. Gallen) for his valuable feedback and advice.
Parts of this work were supported by the BMBF/BMWK projects
XAINES (Grant 01IW20005) and
EuroDaT (Grant 68GX21010K).

\bibliographystyle{ACM-Reference-Format}
\bibliography{library}


\end{document}